\documentclass[letterpaper, 10pt, conference]{IEEEtran}
\IEEEoverridecommandlockouts

\usepackage[T1]{fontenc}
\usepackage{graphicx}
\usepackage{booktabs}
\usepackage{multirow}
\usepackage{amsmath}
\usepackage{amssymb}
\usepackage{textcomp}    
\usepackage{microtype}
\usepackage{stfloats}   
\usepackage[letterpaper, left=57pt, right=57pt, top=56pt, bottom=58pt,
            headheight=0pt, headsep=0pt, footskip=30pt]{geometry}

\usepackage{xcolor}
\usepackage{tikz}
\usetikzlibrary{arrows.meta,positioning,calc,decorations.pathreplacing}
\usepackage[numbers,sort&compress]{natbib}
\usepackage[hidelinks]{hyperref}

\newcommand{\para}[1]{\smallskip\noindent\textbf{#1}\hspace{0.45em}}

\title{\LARGE \bf Latent World Models with Monotone Planning Costs\\for Image-Goal Navigation}

\author{Amirhosein Chahe, Siwei Cai, and Lifeng Zhou
\thanks{The authors are with Drexel University, Philadelphia, PA, USA.
        {\tt\small \{ac4462, sc3568, lz457\}@drexel.edu}}%
}

\begin{document}
\maketitle
\thispagestyle{empty}
\pagestyle{empty}
\bstctlcite{IEEEexample:BSTcontrol}

\begin{abstract}
Image-goal navigation with latent world models requires not only accurate
future prediction but also a planning cost that reliably ranks candidate
action sequences. We use the cosine distance between the predicted future
embedding and the goal embedding as this cost, and show that poor cost
ordering misleads sampling-based planners such as the Cross-Entropy Method
(CEM). We propose a latent world model built on a frozen DINO-family encoder
and train it with two complementary objectives: an autoregressive rollout loss
that closes the gap between training and multi-step planning rollouts, and a
Monotone Cost Ranking (MCR) loss that makes increasingly perturbed action
sequences receive increasingly higher planning costs. Ablations show that this
ordinal shaping outperforms an action-contrastive InfoNCE alternative, which
separates rollouts under different action orderings without ranking them by
deviation. On the GNM navigation dataset our method outperforms Navigation
World Models (NWM), DINO-WM, OmniVLA, and NoMaD, reducing orientation error by
$2.7\times$ over the same-encoder DINO-WM baseline. Deployed zero-shot on a
physical robot, the model follows goal-directed paths in unseen indoor and
outdoor environments.
\end{abstract}

\section{Introduction}
\label{sec:intro}
Image-goal navigation asks a robot to reach a location specified by a
reference image. Beyond matching the current observation to the goal, the
robot must predict how candidate actions will change its future observations
and select the action sequence that moves it toward the target view.

\begin{figure*}[t!]
  \centering
  \resizebox{\textwidth}{!}{%
    \input{teaser}%
  }
  \vspace{-6mm}
  \caption{%
    \textbf{Latent world model with monotone planning cost.}
    \emph{Top:} a frozen DINO encoder $\phi$ maps the current observation $o_1$
    and the goal $o_g$ to patch features $z_1$ and $z_g$ (shown as PCA colour
    maps). A trainable AdaLN ViT predictor $f_\theta$, conditioned on the
    candidate actions $a_1,\ldots,a_H$, rolls the features forward
    autoregressively to $\hat z_H$; the planning cost is the cosine distance
    between $\hat z_H$ and $z_g$. The \textbf{rollout loss} supervises every step
    of this self-conditioned rollout, matching the regime CEM uses at test time,
    and the \textbf{Monotone Cost Ranking (MCR) loss} ranks graded action
    perturbations so that larger trajectory deviation incurs higher cost.
    \emph{Bottom left:} CEM candidates for this start/goal pair, coloured by
    planning cost. \emph{Bottom right:} the cost landscape over
    $(\Delta\mathrm{heading},\Delta\mathrm{forward})$ perturbations of the
    goal-reaching plan. Teacher-forced DINO-WM yields a nearly flat surface with
    no basin at the true plan, so the planner drifts; ours has a clear basin whose
    cost grows monotonically with deviation, and the planner heads for the goal.%
  }
  \label{fig:teaser}
  \vspace{-3mm}
\end{figure*}

Existing learning-based methods follow two paradigms. Reactive policies such
as ViNT~\citep{Shah2023ViNT}, NoMaD~\citep{Sridhar2024NoMaD}, and large VLA
models such as OmniVLA~\citep{Hirose2025OmniVLAAO} map the current observation
and goal directly to actions without simulating future states. World-model
methods instead couple learned dynamics with Model Predictive Control (MPC):
Navigation World Models (NWM)~\citep{Bar2024NavigationWM} plan over predicted
video frames, and DINO-WM~\citep{Zhou2024DINOWM} plans more efficiently over
predicted DINOv2 patch features. DINO-WM, however, is trained with single-step
teacher forcing, whereas MPC rolls the model out autoregressively from its own
predictions. This train--test mismatch accumulates error, and standard
regression losses (MSE, cosine) do not guarantee that the induced cosine
planning cost ranks candidate action sequences correctly.

We argue that a latent world model for MPC must satisfy two requirements
(Fig.~\ref{fig:teaser}): \emph{rollout consistency}, i.e., accurate
predictions when conditioned on its own previous outputs, and a
\emph{monotone cost landscape}, i.e., action sequences that deviate further
from a goal-reaching trajectory receive higher cost. We therefore propose a
JEPA-style latent world model with a frozen DINO-family
encoder~\citep{Baldassarre2025BackTT} and a trainable AdaLN ViT predictor,
trained with (i) an autoregressive rollout loss whose horizon grows through a
counter-curriculum, matching the self-conditioned regime of CEM planning, and
(ii) Monotone Cost Ranking (MCR), a pairwise ranking loss over graded action
perturbations that makes the cosine cost increase with the dead-reckoned
deviation of each perturbed sequence from the ground-truth trajectory.

Shaping this cost raises a design choice. Contrastive objectives are the
standard tool for making a representation action-sensitive: they separate the
rollout under the true actions from rollouts under other action sequences.
CEM, however, is a rank-based optimizer that keeps the lowest-cost candidates,
so it needs the cost to be \emph{ordered} by trajectory deviation, not merely
separable. MCR encodes this ordering directly. Our ablation confirms the
distinction: an action-contrastive InfoNCE term, which pushes every negative
away with equal force regardless of how far its trajectory deviates, degrades
planning, whereas MCR's graded margins consistently improve it.

Trained on the GNM dataset~\citep{Shah2023GNM}, our model outperforms
NWM~\citep{Bar2024NavigationWM}, DINO-WM~\citep{Zhou2024DINOWM},
OmniVLA~\citep{Hirose2025OmniVLAAO}, and NoMaD~\citep{Sridhar2024NoMaD} on
orientation and worst-case displacement metrics, reduces orientation error by
$2.7\times$ over the same-encoder DINO-WM baseline, and plans entirely in
latent space. In summary, our contributions are:
\begin{itemize}
  \item an autoregressive rollout loss with a counter-curriculum that trains
        latent world models in the multi-step self-conditioned regime used by
        CEM;
  \item the principle that the planning cost of a latent world model should be
        monotone in trajectory deviation, and Monotone Cost Ranking (MCR), a
        pairwise margin-ranking objective over graded action perturbations that
        enforces it; an ablation of cost-shaping objectives shows that ranking
        candidates by deviation outperforms separating them contrastively
        (InfoNCE), with gains that hold across two encoders and both
        teacher-forced and rollout-trained predictors;
  \item state-of-the-art results on GNM against world-model and reactive
        baselines, and zero-shot deployment on a physical robot.
\end{itemize}

\section{Related Work}
\label{sec:related}

\para{Visual navigation and world models.}
Learning-based navigation has moved from robot-specific policies to
foundation-style models trained on diverse offline data.
GNM~\citep{Shah2023GNM} introduced cross-embodiment goal-conditioned
navigation, ViNT~\citep{Shah2023ViNT} added a Transformer backbone and
large-scale pretraining, NoMaD~\citep{Sridhar2024NoMaD} unified goal reaching
and exploration with a goal-masked diffusion policy, and
OmniVLA~\citep{Hirose2025OmniVLAAO} scales navigation with multimodal
vision-language-action models. These policies generalize well but are
reactive: they do not simulate the consequences of candidate action
sequences, so they cannot compare long-horizon plans at test time.
World-model methods add this capability. NWM~\citep{Bar2024NavigationWM}
runs MPC over frames generated by a goal-conditioned video diffusion model,
at high computational cost; DINO-WM~\citep{Zhou2024DINOWM} instead predicts
future DINOv2 patch features and plans in latent space;
X-Mobility~\citep{Liu2024XMobility} shows that decoupled latent world models
transfer zero-shot from simulation to the real world. We build on the latent
direction and connect it to continuous SE(2) action-sequence optimization for
physical image-goal navigation.

\para{Latent predictive models for planning.}
Predicting future representations instead of pixels is the premise of the
JEPA family~\citep{Balestriero2025LeJEPAPA}, and frozen DINO features have
proven a strong substrate for latent world models~\citep{Baldassarre2025BackTT}.
DINO-WM is the closest baseline: it scores candidate actions by comparing the
predicted final latent with the goal embedding, but it is trained with
single-step teacher forcing while MPC rolls it out from its own predictions.
Our rollout loss removes this mismatch by training under the same
self-conditioned regime.

\para{Training objectives for planning-ready world models.}
Teacher forcing stabilizes sequence training but causes exposure
bias~\citep{Bengio2015ScheduledSampling}; our rollout loss applies the
standard remedy, supervising multi-step predictions conditioned on the
model's own outputs, to latent navigation. A second requirement is a
well-shaped planning objective: if the cosine cost does not increase as
trajectories deviate from the goal-reaching path, CEM can select poor plans
even when the predictor is locally accurate. MCR shapes this cost directly by
ranking perturbed action sequences by their trajectory deviation. Contrastive
objectives~\citep{vandenOord2018CPC,Sermanet2018TCN,Laskin2020CURL,Eysenbach2022CRL}
shape representations by separating positives from negatives; MCR instead
imposes an \emph{ordering} by deviation magnitude, which our ablation
(\S\ref{sec:experiments:ablation}) shows is what a rank-based planner needs.
Further related work is discussed in
Appendix~\ref{app:extended_related}.

\section{Method}
\label{sec:method}

\begin{figure*}[!b]
    \centering
    \vspace{-2mm}
    \begin{tabular}{@{}c@{\hspace{0.01\textwidth}}c@{\hspace{0.015\textwidth}}c@{}}
        \includegraphics[width=0.15\textwidth]{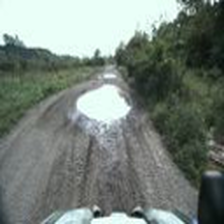} &
        \includegraphics[width=0.15\textwidth]{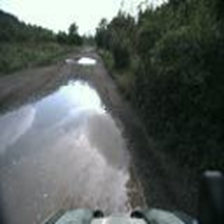} &
        \includegraphics[width=0.5\textwidth]{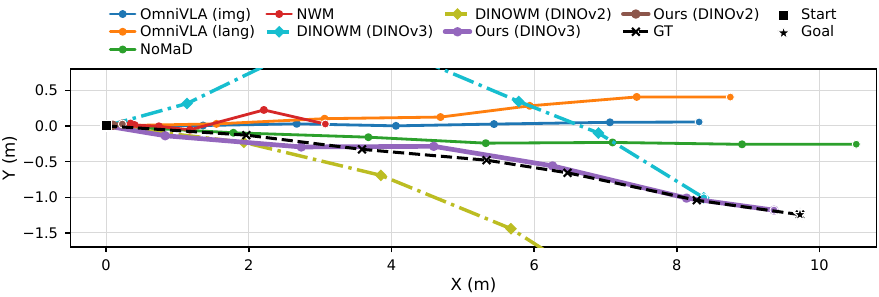} \\
        {\footnotesize (a) Start $o_1$} & {\footnotesize (b) Goal $o_g$} & {\footnotesize (c) Planned trajectories in the robot frame}
    \end{tabular}
    \vspace{-2mm}
    \caption{Qualitative example on a held-out GNM episode. From the start (a)
    and goal (b) observations, each method plans a trajectory (c). Our MCR
    model tracks the ground-truth path most closely; several baselines
    under-progress or deviate.}
    \label{fig:trajectory_example_255}
\end{figure*}

\subsection{Problem Formulation}
\label{sec:method:problem}
Given a short history of RGB observations and a goal image, we seek an action
sequence that drives the robot to the depicted location. Each action is a
continuous body-frame SE(2) displacement with the yaw change encoded as a
sine--cosine pair, $(\Delta x,\Delta y,\sin\Delta\theta,\cos\Delta\theta)$,
which avoids angular discontinuities. No map, depth, or GPS is used; odometry
provides action labels during training only.

\subsection{Architecture}
\label{sec:method:arch}
\para{Frozen DINO encoder.}
A frozen DINO-family encoder maps each observation $O_t$ to a spatial grid of
patch features $\mathbf{z}_t=\phi(O_t)$, layer-normalized as
$\tilde{\mathbf{z}}_t=\mathrm{LN}(\mathbf{z}_t)$. Only the predictor is
trained.

\para{AdaLN ViT predictor.}
The predictor $f_\theta$ is a causal Vision Transformer with rotary
positional encodings and Adaptive Layer Normalization
(AdaLN)~\citep{Peebles2022DiT}. Given a context window of $C$ recent feature
maps and actions it predicts the next feature map,
\begin{equation}
  \hat{\mathbf{z}}_{t+1} = f_\theta\!\left(
      \tilde{\mathbf{z}}_{t-C+1:t},\;\mathbf{a}_{t-C+1:t}
  \right).
  \label{eq:predictor}
\end{equation}
A small MLP maps each action to per-layer scale-and-shift parameters that
modulate every block, unlike DINO-WM~\citep{Zhou2024DINOWM}, which
concatenates a flat action token onto each patch. Applying $f_\theta$
recursively, appending each prediction to the context, yields the $H$-step
rollout operator
\begin{equation}
  F_\theta^{H}\!\left(\tilde{\mathbf{Z}}_{\mathrm{ctx}},\,
    \mathbf{a}_{1:H}\right)
  \;\mapsto\;\hat{\mathbf{z}}_{H}.
  \label{eq:rollout_op}
\end{equation}

\subsection{Training Objectives}
\label{sec:method:objectives}
Training has a \emph{base phase} with autoregressive rollout supervision and
a brief \emph{MCR phase} that shapes the cost landscape.

\subsubsection{Rollout Loss}
\label{sec:method:rollout}
The per-step prediction loss combines L2 and cosine terms,
\begin{equation}
  \mathcal{L}_{\mathrm{pred}}(\hat{\mathbf{z}},\mathbf{z})
  = \|\hat{\mathbf{z}}-\mathbf{z}\|_2^2
  + \lambda_{\mathrm{cos}}\,\bigl(1-\cos(\hat{\mathbf{z}},\mathbf{z})\bigr),
  \label{eq:pred_loss}
\end{equation}
with $\lambda_{\mathrm{cos}}\!=\!0.5$. Instead of teacher forcing, we unroll
the predictor for $K$ steps from a length-$S$ ground-truth seed and condition
each step on the rolling buffer
$\mathbf{u}^{(k)}_{j}=\tilde{\mathbf{z}}_{j}$ for $j\le S$ and
$\mathbf{u}^{(k)}_{j}=\hat{\mathbf{z}}^{(j-S)}$ for $j>S$:
\begin{align}
  \hat{\mathbf{z}}^{(k)}
  &= f_\theta\!\left(
        \mathbf{u}^{(k)}_{S+k-C:\,S+k-1},\;
        \mathbf{a}_{S+k-C:\,S+k-1}
      \right), \label{eq:rollout_step}\\[1pt]
  \mathcal{L}_{\mathrm{rollout}}
  &= \frac{1}{K}\sum_{k=1}^{K}
       \mathcal{L}_{\mathrm{pred}}\!\left(
         \hat{\mathbf{z}}^{(k)},\;
         \tilde{\mathbf{z}}_{S+k}
       \right).
  \label{eq:rollout}
\end{align}
This mirrors the MPC rollout at test time. To avoid collapse at long
horizons, a \emph{counter-curriculum} ramps $K$ from 2 to 8 during training
while annealing the learning rate and batch size
(Appendix~\ref{app:training}).

\subsubsection{Monotone Cost Ranking (MCR)}
\label{sec:method:surface}
CEM is a rank-based optimizer: at every iteration it keeps the lowest-cost
candidates and refits its sampling distribution to them, so what it requires
from the cost is a correct \emph{ordering} of candidates by how far they
deviate from the goal-reaching trajectory. Even after rollout training, the
cosine cost between the predicted final embedding and the goal embedding need
not be monotone in this deviation, and local minima can trap the planner. MCR
enforces this monotonicity directly.
We perturb the ground-truth action sequence with AR(1)-correlated Gaussian
noise ($\rho\!=\!0.7$) at three levels $\sigma\!\in\!\{0.05,0.15,0.35\}$. For
each candidate $i$ we compute its cosine cost
$c_i = 1-\cos\!\bigl(F_\theta^{H}(\tilde{\mathbf{Z}}_{\mathrm{ctx}},
\mathbf{a}^{(i)}_{1:H}),\,\tilde{\mathbf{z}}_g\bigr)$ and its trajectory
distance $d_i$, the mean L2 deviation of its dead-reckoned path from the
ground-truth path. A pairwise margin-ranking loss requires candidates farther
in trajectory space to incur proportionally higher cost:
\begin{equation}
  \mathcal{L}_{\mathrm{mcr}}
  = \frac{1}{|\mathcal{P}|}
    \sum_{(i,j)\in\mathcal{P}}
    \max\!\bigl(0,\;m_{ij}+c_j-c_i\bigr),
  \label{eq:mcr}
\end{equation}
where $\mathcal{P}$ contains ordered pairs with $d_i>d_j$ and
$m_{ij}=\max(0.02,\,0.5\,|d_i-d_j|)$. The MCR phase minimizes
$\mathcal{L}_{\mathrm{MCR}}=\mathcal{L}_{\mathrm{rollout}}+\mathcal{L}_{\mathrm{mcr}}$,
where the rollout term on the ground-truth sequence anchors prediction
quality.

\begin{figure}[t]
    \centering
    \includegraphics[width=\linewidth]{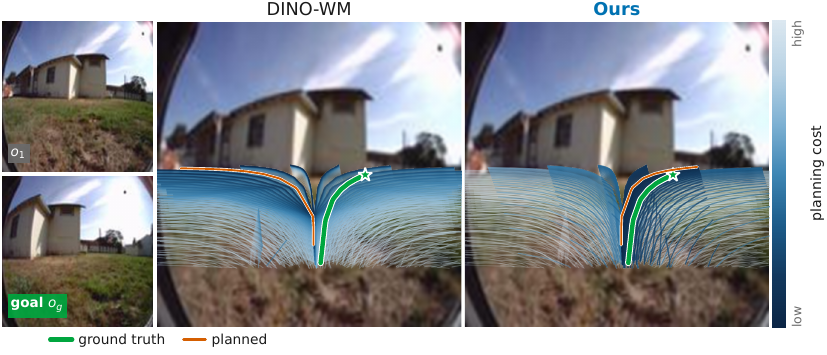}\\[3pt]
    \includegraphics[width=\linewidth]{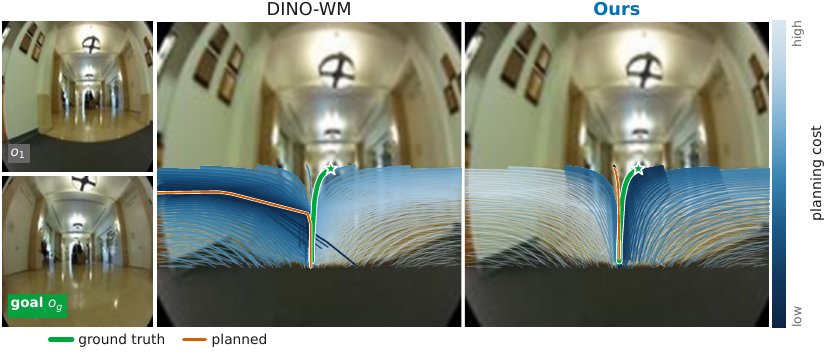}
    \vspace{-8mm}
    \caption{\textbf{CEM candidates on held-out GNM episodes} (outdoor, top;
    indoor, bottom). For each episode (start $o_1$ and goal $o_g$ at left),
    the sampled CEM action sequences are dead-reckoned and drawn on the
    current view, coloured by planning cost (dark = low); green is the
    ground-truth trajectory, orange the plan, and the star the goal position.
    }
    \label{fig:gnm_candidates}
    \vspace{-3mm}
\end{figure}

\subsubsection{Alternative: Contrastive Cost Shaping}
\label{sec:method:contrastive}
A natural alternative to ranking is to make the rollout action-sensitive by
contrast: an action-contrastive InfoNCE term (ACT) that separates the rollout
under the true actions from rollouts of the \emph{same scene} under other
orderings of those actions. Let $\mathrm{pool}(\cdot)$ denote
$\ell_2$-normalised spatial mean-pooling, $\mathbf{g}^{(i)}$ the pooled
ground-truth future features of batch element $i$,
$\mathbf{p}^{(i)}_{+}=\mathrm{pool}\bigl(F_\theta^{K}(\tilde{\mathbf{Z}}^{(i)}_{\mathrm{ctx}},\mathbf{a}^{(i)}_{1:K})\bigr)$
the pooled positive rollout, and
$\mathbf{p}^{(i)}_{n}=\mathrm{pool}\bigl(F_\theta^{K}(\tilde{\mathbf{Z}}^{(i)}_{\mathrm{ctx}},\pi_n(\mathbf{a}^{(i)}_{1:K}))\bigr)$
the rollout under a reordering $\pi_n$ of the same sequence. Each element has
$N\!=\!4$ negatives (three random non-identity permutations and the reversal),
so the contrast isolates action order:
\begin{equation}
  \mathcal{L}_{\mathrm{act}}
  = -\frac{1}{B}\sum_{i}\log
    \frac{e^{\mathbf{g}^{(i)\top}\mathbf{p}^{(i)}_{+}/\tau}}
         {e^{\mathbf{g}^{(i)\top}\mathbf{p}^{(i)}_{+}/\tau}
          +\sum_{n=1}^{N}e^{\mathbf{g}^{(i)\top}\mathbf{p}^{(i)}_{n}/\tau}}.
  \label{eq:act}
\end{equation}
ACT is added to the rollout loss during the base phase,
$\mathcal{L}=\mathcal{L}_{\mathrm{rollout}}+\lambda_{\mathrm{act}}\mathcal{L}_{\mathrm{act}}$
(Appendix~\ref{app:act_hparams}). The two objectives differ in what they
enforce: InfoNCE pushes every negative away from the positive with the same
force regardless of how far its trajectory deviates, whereas the MCR margin
$m_{ij}$ scales with $|d_i-d_j|$, so nearly identical candidates are
constrained only weakly. Section~\ref{sec:experiments:ablation} compares the
two.

\subsection{CEM Planning at Inference}
\label{sec:method:cem}
Given the recent observations and a goal image, we plan with the
Cross-Entropy Method. Candidate action sequences are sampled from a unicycle
kinematic prior, converted to SE(2) actions, and scored by rolling out the
world model in latent space:
\begin{equation}
  \mathbf{A}^* = \arg\min_{\mathbf{A}}\;
  1 - \mathrm{sim}\!\left(
    \mathrm{pool}\!\left(
      F_\theta^{H}(\tilde{\mathbf{Z}}_{\mathrm{ctx}},\,\mathbf{A})
    \right),\;
    \mathrm{pool}(\tilde{\mathbf{z}}_g)
  \right).
  \label{eq:cem}
\end{equation}
CEM refits a Gaussian to the top-$k$ elites for a fixed number of iterations;
the first action of the best sequence is executed and the robot replans at
every step. No pixel reconstruction is needed.

\section{Experiments}
\label{sec:experiments}

\subsection{Setup}
\label{sec:experiments:setup}
\para{Dataset.}
We train and evaluate on GNM~\citep{Shah2023GNM}, a cross-embodiment corpus
of 60+ hours from six robot platforms, using a held-out split of 1,316
trajectories. CAST~\citep{Glossop2025CASTCL} adds language annotations to
GNM; our model uses images only, while the language-conditioned OmniVLA
baseline uses CAST instructions.

\para{Implementation.}
We evaluate DINOv2-ViT-S/14 and DINOv3-ViT-S/16 as frozen encoders. The
predictor has 6 layers, 12 heads, embedding dimension 384, RoPE, and a causal
mask. Training uses 4$\times$NVIDIA H200 GPUs; MCR is a 1,000-step phase on
top of the converged Base checkpoint. Full optimizer settings, step counts,
and the counter-curriculum schedule are in Appendix~\ref{app:training}.
Real-world inference runs on an NVIDIA Jetson AGX Orin with an 8-step
horizon.

\para{Protocol and metrics.}
Our model and DINO-WM plan with CEM ($S\!=\!4$ context frames, $K\!=\!6$
horizon; Appendix~\ref{app:CEM-Hyper}); NWM uses the same CEM optimizer over
its video rollouts; reactive policies use their original inference. Over the 6-step trajectory we report average and worst-case
orientation error (\textbf{AOE}/\textbf{MAOE}, \textdegree,
following~\citealp{Liu2024CityWalkerLE}) and average and worst-case
displacement error (\textbf{ADE}/\textbf{MADE}, m). Lower is better.

\para{Baselines.}
\textbf{DINO-WM}~\citep{Zhou2024DINOWM}: our re-implementation with a ViT
predictor, per-patch action concatenation, single-step teacher forcing, and
L2 loss, with both encoders. \textbf{NWM}~\citep{Bar2024NavigationWM}:
pixel-space video-diffusion world model with MPC.
\textbf{OmniVLA}~\citep{Hirose2025OmniVLAAO}: large VLA in image-goal (img)
and language (lang) modes. \textbf{NoMaD (GC)}~\citep{Sridhar2024NoMaD}:
goal-conditioned diffusion policy.

\begin{table}[t]
\centering
\caption{GNM test results ($K\!=\!6$, $C\!=\!4$); lower is better.
\textbf{Bold}: best world-model planner; \underline{underline}: best overall.
TF: teacher-forced; Base: rollout; MCR: Base + monotone cost ranking;
ACT: Base + action-contrastive InfoNCE (alternative cost-shaping objective,
\S\ref{sec:experiments:ablation}).}
\label{tab:main}
\setlength{\tabcolsep}{4pt}
\setlength{\aboverulesep}{0pt}
\setlength{\belowrulesep}{0pt}
\renewcommand{\arraystretch}{1.04}
\footnotesize
\begin{tabular}{@{}l cccc@{}}
\toprule
Method & AOE\,(\textdegree) & MAOE\,(\textdegree) & ADE\,(m) & MADE\,(m)\\
\midrule
\multicolumn{5}{@{}l}{\emph{Reactive policies}}\\
OmniVLA (img)~\citep{Hirose2025OmniVLAAO}  & 7.73  & 11.58 & 0.94 & 1.72 \\
OmniVLA (lang)~\citep{Hirose2025OmniVLAAO} & 10.94 & 17.31 & 1.29 & 2.46 \\
NoMaD (GC)~\citep{Sridhar2024NoMaD}        & 9.09  & 14.20 & \underline{0.88} & 1.67 \\
\midrule
\multicolumn{5}{@{}l}{\emph{World-model planner, pixel space}}\\
NWM~\citep{Bar2024NavigationWM}            & 14.57 & 26.57 & 1.04 & 1.72 \\
\midrule
\multicolumn{5}{@{}l}{\emph{World-model planners, latent space (baseline)}}\\
DINO-WM (DINOv2)~\citep{Zhou2024DINOWM}    & 20.27 & 30.14 & 3.47 & 6.25 \\
DINO-WM (DINOv3)~\citep{Zhou2024DINOWM}    & 21.82 & 32.34 & 2.09 & 3.69 \\
\midrule
\multicolumn{5}{@{}l}{\emph{Ours (DINOv3)}}\\
TF          & 10.52 & 15.29 & 1.26 & 2.10 \\
TF + MCR    &  8.84 & 12.91 & 1.16 & 1.93 \\
Base        & 10.15 & 14.62 & 1.22 & 1.97 \\
ACT         & 21.17 & 29.44 & 1.95 & 3.36 \\
ACT + MCR   & 11.92 & 18.74 & 1.66 & 2.85 \\
MCR         &  8.54 & 12.22 & 1.17 & 1.91 \\
\midrule
\multicolumn{5}{@{}l}{\emph{Ours (DINOv2)}}\\
TF          & 12.10 & 18.55 & 3.28 & 5.57 \\
TF + MCR    & 10.66 & 16.51 & 3.23 & 5.44 \\
Base        &  9.12 & 13.29 & 1.12 & 1.83 \\
ACT         & 13.80 & 21.08 & 1.58 & 2.69 \\
ACT + MCR   & 10.33 & 15.50 & 1.41 & 2.31 \\
\textbf{MCR} & \underline{\textbf{7.63}} & \underline{\textbf{11.11}} & \textbf{0.99} & \underline{\textbf{1.64}} \\
\bottomrule
\end{tabular}
\vspace{-3mm}
\end{table}

\subsection{Main Results}
\label{sec:experiments:main}
Table~\ref{tab:main} reports results on the GNM test split and
Fig.~\ref{fig:trajectory_example_255} shows a representative episode.

\para{Latent-space world models.}
With the DINOv2 encoder, our full recipe reduces AOE by $\mathbf{2.7\times}$
relative to the same-encoder DINO-WM baseline, with similar or larger gains on
displacement; DINOv3 shows the same pattern. The ablations
(\S\ref{sec:experiments:ablation}) attribute this to three compounding
factors: the AdaLN action-conditioned predictor, autoregressive rollout
training, and monotone cost shaping.

\para{Pixel-space planner.}
NWM's AOE is nearly twice ours, so latent planning with a shaped cost surface
outperforms pixel-space video-diffusion MPC without any pixel reconstruction.

\para{Reactive policies.}
MCR (DINOv2) achieves the lowest AOE, MAOE, and MADE of \emph{all} methods,
edging the strongest reactive baselines (OmniVLA, NoMaD); the sole exception
is ADE, where those policies lead by a small margin. Reactive policies imitate
ground-truth trajectories, which helps when test motion follows the training
distribution, whereas our model optimizes goal proximity in embedding space
and is free to reach the goal by a different path
(Fig.~\ref{fig:trajectory_example_255}). That a latent world model matches or
exceeds reactive policies on three of four metrics, without imitation, is the
central finding of this work.

\subsection{Ablation Study}
\label{sec:experiments:ablation}
\begin{figure*}[t]
    \centering
    \includegraphics[width=0.85\textwidth]{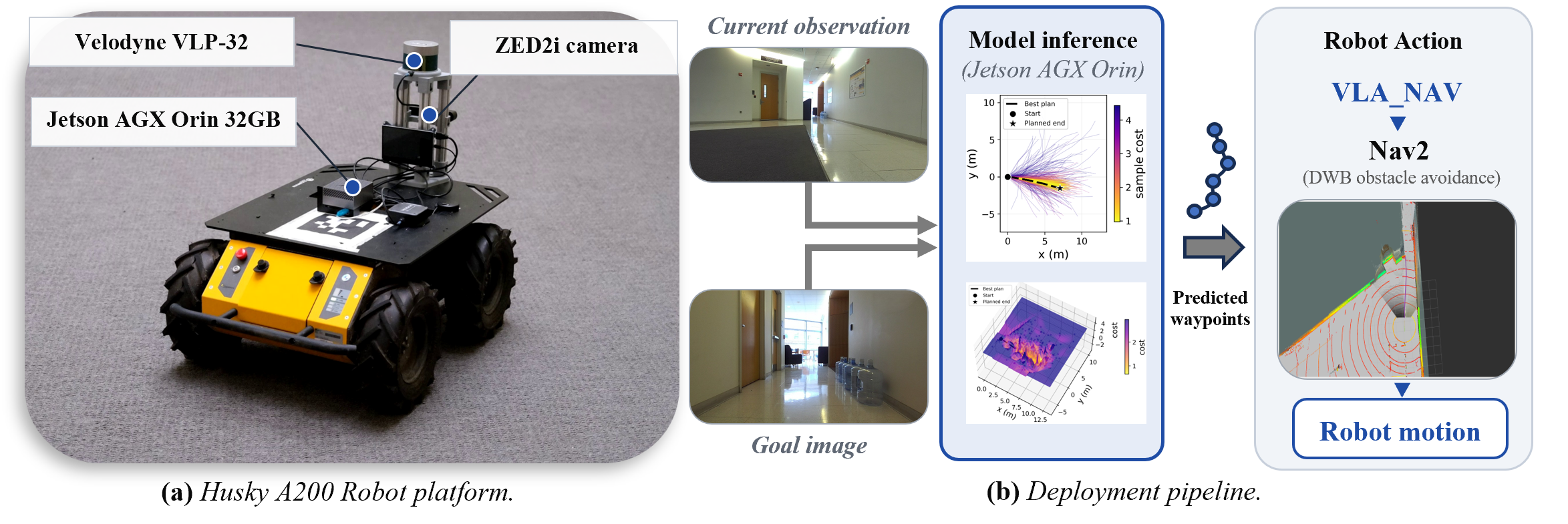}
    \vspace{-2mm}
    \caption{Real-world robot platform and deployment pipeline.
    (a) Clearpath Husky A200 equipped with a Velodyne VLP-32 lidar, a ZED2i
    stereo camera, and an NVIDIA Jetson AGX Orin 32GB onboard computer.
    (b) Given the current observation and goal image, the model performs
    onboard inference and predicts navigation waypoints, which are passed
    through \texttt{vla\_nav} to the ROS~2 Nav2 stack, where DWB-based local
    obstacle avoidance generates executable robot motion.}
    \label{fig:real_robot_pipeline}
    \vspace{-4mm}
\end{figure*}
\begin{figure*}[!b]
    \centering
    \includegraphics[width=0.9\textwidth]{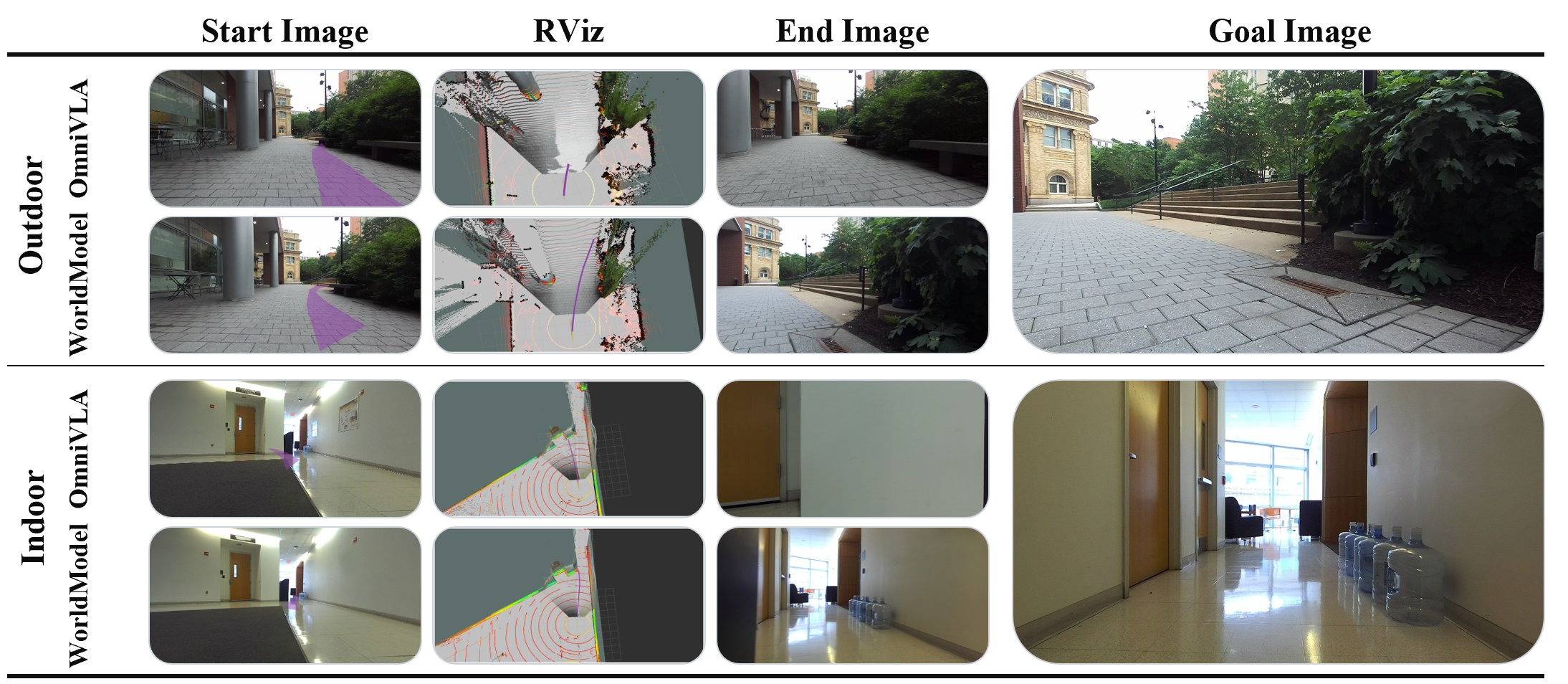}
    \vspace{-4mm}
    \caption{Zero-shot real-world rollouts in outdoor and indoor scenes. For
    each scenario, the top row is OmniVLA and the bottom row is ours. Neither
    method is trained or fine-tuned on the deployment scenes or on data from
    the Husky platform. OmniVLA produces overly short trajectories or heads
    toward the wall, whereas our model follows feasible goal-directed paths
    and ends closer to the target viewpoint.}
    \label{fig:qualitative_nav_layout}
\end{figure*}
\para{Rollout training vs.\ teacher forcing.}
Autoregressive rollout beats single-step teacher forcing before and after MCR
on both encoders (Table~\ref{tab:main}); the gap is largest for DINOv2
(9.12\textdegree{} vs.\ 12.10\textdegree{} AOE before MCR). Matching the
training distribution to the self-conditioned rollouts used by CEM directly
improves multi-step planning.

\para{Effect of Monotone Cost Ranking.}
MCR gives the largest and most consistent per-stage improvement, reducing AOE
for every base variant and both encoders. Fig.~\ref{fig:gnm_candidates}
shows the mechanism: the DINO-WM cost varies across CEM candidates but has no
basin around the ground-truth trajectory, so its low-cost elites do not point
toward the goal and the resulting plan is wrong, whereas our cost forms a
narrow low-cost basin around the ground-truth trajectory that the planner
follows.

\para{Encoder choice.}
DINOv2-ViT-S/14 is the best encoder under the MCR recipe (7.63\textdegree{}
vs.\ 8.54\textdegree{} for DINOv3-ViT-S/16), but the ordering of training
variants is identical for both, so the benefits of rollout training and MCR
are not tied to a specific backbone.

\para{Cost shaping: ranking vs.\ contrastive separation.}
MCR and ACT both shape the cost from rollouts of the same scene under
alternative action sequences, but they enforce different properties. ACT
(\S\ref{sec:method:contrastive}) separates the true rollout from rollouts under
reorderings of the same actions, whereas MCR orders candidates by trajectory
deviation, which is what the planner consumes. ACT raises AOE on both
encoders, with DINOv3 reaching the level of the DINO-WM baseline, whereas MCR
lowers it. Adding MCR on top of ACT recovers much of the loss but remains well
behind MCR alone, so the contrastive term is harmful rather than redundant,
even though ACT is trained jointly for the full 18k-step base schedule while
MCR is a 1,000-step phase. We attribute this to the inductive bias of the two
objectives: permutation negatives share the positive's action multiset, so
their dead-reckoned paths are nearly identical, yet InfoNCE separates them as
strongly as any other negative and thereby distorts the latent geometry that
accurate rollout and the cosine cost rely on. The MCR margin scales with the
deviation gap, so such close candidates are constrained only weakly.
Hyperparameters and further analysis are in
Appendices~\ref{app:act_hparams} and~\ref{app:act_analysis}.

\subsection{Real-World Robot Demonstration}
\label{sec:experiments:realworld}
We further demonstrate the method on a physical robot to verify that the
predicted actions can be executed through a real navigation stack. All
methods are evaluated zero-shot: neither OmniVLA nor our model is trained or
fine-tuned on the deployment environments or on data from this platform.

\para{Robot platform.}
As shown in Fig.~\ref{fig:real_robot_pipeline}(a), the demonstration uses a
Clearpath Husky A200 unmanned ground vehicle running ROS~2 Humble with an
NVIDIA Jetson AGX Orin 32GB module for onboard computation, a ZED2i stereo
camera for visual observations, and a Velodyne VLP-32 lidar for geometric
sensing and local navigation. The learned model predicts waypoint-level
actions from RGB alone; the lidar-based navigation stack provides the
low-level feasibility checks and obstacle avoidance needed for safe physical
execution.

\para{Deployment pipeline.}
Fig.~\ref{fig:real_robot_pipeline}(b) shows how the model is connected to
the execution stack. At each replanning step the system receives the current
camera observation and a goal image; the model runs onboard the Jetson and
predicts a sequence of navigation waypoints. A custom \texttt{vla\_nav}
package converts them into Nav2 goals, which are executed with the NavFn
global planner (Dijkstra search), the DWB local controller, and the BT
Navigator. This separates image-goal waypoint prediction from collision
avoidance, velocity command generation, and short-horizon feasibility. If a
predicted waypoint is reachable, the robot follows the Nav2 trajectory until
the goal is reached or the next replanning step is triggered. If it is blocked
or infeasible, the executor applies a progressive sub-goal fallback, trying
shortened waypoints along the same direction at 75\%, 50\%, 25\%, and 10\% of
the original distance before declaring the waypoint infeasible.

\para{Qualitative results.}
Fig.~\ref{fig:qualitative_nav_layout} shows representative rollouts in
outdoor and indoor environments. Outdoors, OmniVLA generates an overly short
trajectory and terminates before making sufficient progress; indoors it moves
toward the wall instead of continuing along the free corridor. In both scenes
our model follows feasible free-space paths and reaches final observations
that closely match the target views, showing stronger spatial coherence and
more reliable waypoint prediction under zero-shot deployment.

\para{Trial outcomes.}
Across five representative test scenarios, our model reached the target in
all five trials, whereas OmniVLA succeeded in two of five. A trial is counted
as successful if the robot reaches a final observation visually consistent
with the goal view without collision or manual intervention. This small
number of trials is not a statistical benchmark, but it supports that the
latent world-model planner transfers more reliably to physical deployment
than the reactive baseline in these indoor and outdoor settings.

\section{Conclusion and Limitations}
\label{sec:conclusion}
We presented a latent world model for image-goal navigation built around a
\emph{monotone planning cost}: the cosine distance between predicted and goal
embeddings should rise with trajectory deviation so that gradient-free CEM
converges reliably. An autoregressive rollout loss closes the teacher-forcing
train/test gap and a margin-ranking objective shapes the cost landscape.
Together they reduce orientation error by $2.7\times$ over the same-encoder
DINO-WM baseline and outperform all evaluated baselines, including reactive
policies, on orientation and worst-case displacement, entirely in the latent
space of a frozen encoder. Our ablation of cost-shaping objectives shows that
the cost should be shaped by \emph{ordering} candidates by their trajectory
deviation rather than by contrastively separating them: graded ranking
improves planning consistently, whereas an action-contrastive InfoNCE term
does not.

Our evaluation concentrates on static or low-traffic environments; dynamic,
multi-agent scenes remain untested, and the cost, shaped purely by goal
proximity in embedding space, does not model other agents. The autoregressive
predictor also accumulates error over the horizon; the rollout loss reduces
this mismatch, but very long horizons remain challenging.

\bibliographystyle{IEEEtranN}
\begingroup
\emergencystretch=1.5em
\bibliography{references,bstctl}
\endgroup

\clearpage
\makeatletter
\setlength{\@fptop}{0pt}
\setlength{\@dblfptop}{0pt}
\makeatother
\appendices

\section{Extended Related Work}
\label{app:extended_related}

We provide additional discussion of related visual navigation, latent world-model, contrastive learning, and energy-based planning methods that are complementary to the main related work.
Recent visual navigation methods have explored image-goal primitives, structured trajectory priors, flow- or diffusion-based action generation, socially aware navigation, data scaling, and reinforcement-learning-enhanced navigation policies~\citep{Chen2024IGORIR,Luo2026StepNavST,Gode2024FlowNavCF,Chen2025SocialNavTH,Suomela2026DataSF,He2025FromST}.
Beyond navigation, latent predictive models have been applied to reward-free planning, manipulation, video prediction, latent action learning, and hierarchical planning~\citep{Sobal2025LearningFR,Terver2025WhatDS,Assran2025VJEPA2,Sun2026VLAJEPAEV,Garrido2026LearningLA,Maes2026LeWorldModelSE,Zhang2026HWM,Huang2025Vid2WorldCV,Hou2026WMSurveyRobot}.
Our training objectives are also related to scheduled sampling, DAgger, latent rollout training, contrastive predictive learning, and energy- or value-based goal-reaching objectives~\citep{Bengio2015ScheduledSampling,Ross2011DAgger,Hafner2023DreamerV3,vandenOord2018CPC,Sermanet2018TCN,Chen2020SimCLR,Laskin2020CURL,Eysenbach2022CRL,Florence2021IBC,Ma2022VIPRL,Wang2025TargetBenchCW}.

\section{Training Details}
\label{app:training}

\subsection{Training Configuration Summary}

Table~\ref{tab:configs} summarises the loss objectives and key hyperparameters
for each training variant evaluated in this work.
All variants share the same frozen DINO-family encoder, 6-layer AdaLN ViT
predictor (embedding dim 384, 12 heads, RoPE, causal mask), AdamW optimiser
($\beta_1\!=\!0.9$, $\beta_2\!=\!0.999$, weight decay 0.01, grad clip 1.0),
bfloat16 mixed precision, and 4$\times$H200 GPUs.

\begin{table}[ht]
\centering
\caption{Per-variant training configurations.
Batch sizes are per-GPU; $\to$ denotes values that change with the
rollout counter-curriculum (Table~\ref{tab:schedule_app}).
MCR resumes from the converged Base checkpoint.}
\label{tab:configs}
\setlength{\tabcolsep}{3pt}
\footnotesize
\begin{tabular}{@{}llcccc@{}}
\toprule
Variant & Objective & $K$ & Steps & LR & Batch \\
\midrule
TF
  & $\mathcal{L}_{\mathrm{rollout}}$ ($K\!=\!1$)
  & 1 & 20k & $5\!\times\!10^{-4}$ & 64 \\
Base
  & $\mathcal{L}_{\mathrm{rollout}}$
  & 2$\to$8 & 18k & $10^{-4}\!\to\!5\!\times\!10^{-6}$ & 128$\to$32 \\
ACT
  & $\mathcal{L}_{\mathrm{rollout}} + 0.25\,\mathcal{L}_{\mathrm{act}}$
  & 2$\to$8 & 18k & $10^{-4}\!\to\!5\!\times\!10^{-6}$ & 128$\to$32 \\
MCR
  & $\mathcal{L}_{\mathrm{rollout}} + \mathcal{L}_{\mathrm{mcr}}$
  & 8 & +1k & $10^{-5}$ & 8 \\
\bottomrule
\end{tabular}
\end{table}

\subsection{Base Rollout Counter-Curriculum}

Base (and ACT) training uses a counter-curriculum that ramps the
autoregressive rollout horizon $K$ from 2 to 8 steps, simultaneously
annealing the learning rate and reducing batch size to accommodate the
increased compute per step.
Table~\ref{tab:schedule_app} gives the full schedule.

\begin{table}[ht]
\centering
\caption{Rollout counter-curriculum schedule (Base and ACT variants).
All batch sizes are per-GPU (4$\times$H200 GPUs).}
\label{tab:schedule_app}
\setlength{\tabcolsep}{7pt}
\footnotesize
\begin{tabular}{cccc}
\toprule
Training steps & Rollout $K$ & Batch/GPU & Learning rate \\
\midrule
0--2k    & 2 & 128 & $1\!\times\!10^{-4}$ \\
2k--6k   & 4 & 64  & $5\!\times\!10^{-5}$ \\
6k--10k  & 6 & 32  & $1\!\times\!10^{-5}$ \\
10k--18k & 8 & 32  & $5\!\times\!10^{-6}$ \\
\bottomrule
\end{tabular}
\end{table}

\subsection{MCR Hyperparameters}

MCR resumes from the converged Base checkpoint and runs for 1,000 steps at a
fixed LR of $10^{-5}$ (no warm-up, batch size 8 per GPU).
The rollout horizon is fixed at $K\!=\!8$ to match the last phase of Base
training.
Perturbed action sequences are generated at three noise levels
$\sigma\!\in\!\{0.05,\,0.15,\,0.35\}$ with 1 sample per level (3 perturbed
candidates plus the GT anchor per batch element).
Noise is AR(1)-correlated with $\rho\!=\!0.7$.
The ranking margin is $m_{ij}=\max(0.02,\;0.5\,|d_i-d_j|)$; the GT anchor
loss weight is $\lambda_{\mathrm{anchor}}\!=\!1.0$.

\subsection{ACT Hyperparameters}
\label{app:act_hparams}
ACT (Eq.~\eqref{eq:act}) uses temperature $\tau\!=\!0.1$ and is added to
$\mathcal{L}_{\mathrm{rollout}}$ with weight $\lambda_{\mathrm{act}}\!=\!0.25$
after a 100-step warm-up, under the same rollout counter-curriculum as Base
(Table~\ref{tab:schedule_app}). Each batch element contributes $N\!=\!4$
negatives: three random permutations of its $K$ driving actions, with the
identity permutation rejected, plus the reversed sequence. No cross-scene
negatives are used.

\subsection{CEM Planning Hyperparameters}
\label{app:CEM-Hyper}
CEM uses a planning horizon of $K\!=\!6$ steps and a context window of
$S\!=\!4$ seed frames.
Each iteration samples 128 candidate action sequences from a unicycle
kinematic model with AR(1)-correlated noise ($\rho\!=\!0.5$) for temporal
smoothness; the Gaussian is refit to the top 4 elites, and the search runs
for 6 iterations.

\section{Additional Analysis of Cost-Shaping Objectives}
\label{app:act_analysis}

\para{Mechanism.}
The negatives in Eq.~\eqref{eq:act} are reorderings of the same action
multiset as the positive, so positive and negative rollouts share nearly the
same aggregate displacement but must map to distinct embeddings because their
temporal order differs. Separating such sequences with InfoNCE introduces
contrastive pressure that distorts the local geometry of the predictor's
output space, which CEM relies on when it treats cosine distance in that space
as a planning cost. MCR can partially repair the induced cost surface
(ACT$+$MCR in Table~\ref{tab:main}), but it does not restore the prediction
accuracy lost through this distortion.

\para{Scope.}
The pattern is the same for the DINOv2 and DINOv3 encoders. It reflects a
tension between two objectives: rollout training rewards geometric fidelity to
the true future embedding, whereas order-sensitive contrastive training
rewards separation between embeddings produced by different orderings of the
same actions. Ranking by deviation, as in MCR, asks only for a small margin
between such nearby candidates and therefore does not conflict with rollout
fidelity. If a contrastive term is nonetheless desired, a separate projection
head, gradient surgery, alternating objectives, or applying the loss to a
detached copy of the embedding could reduce the conflict.

\section{Additional CEM Candidate Visualizations}
\label{app:more_examples}

\begin{figure}[!t]
    \centering
    \includegraphics[width=\linewidth]{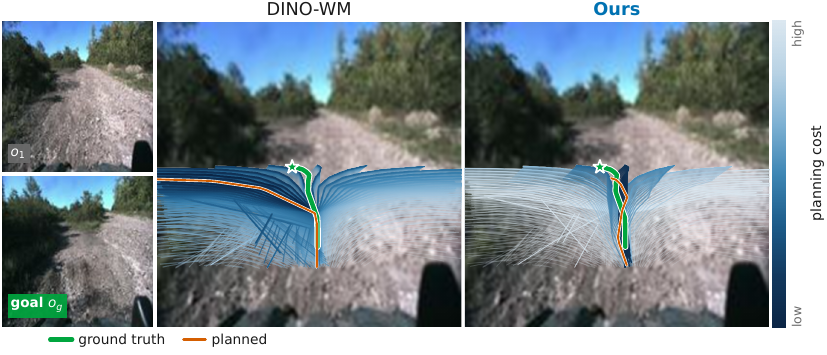}\\[3pt]
    \includegraphics[width=\linewidth]{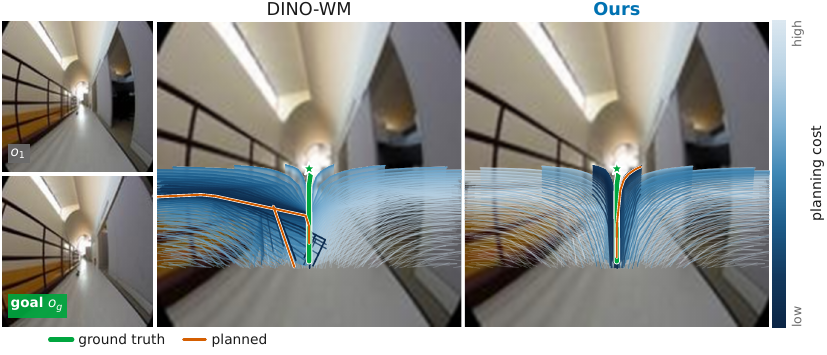}\\[3pt]
    \includegraphics[width=\linewidth]{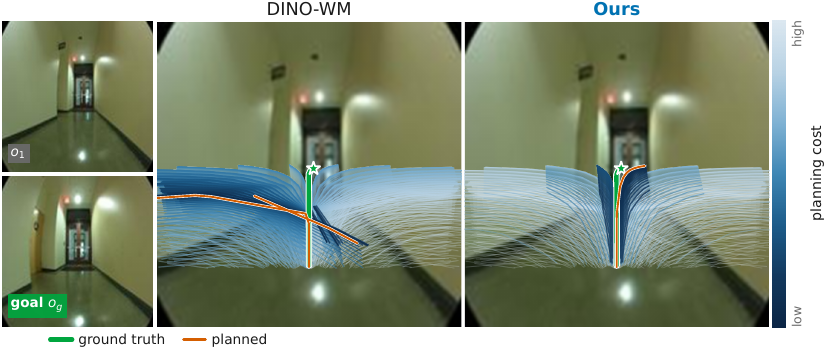}\\[3pt]
    \includegraphics[width=\linewidth]{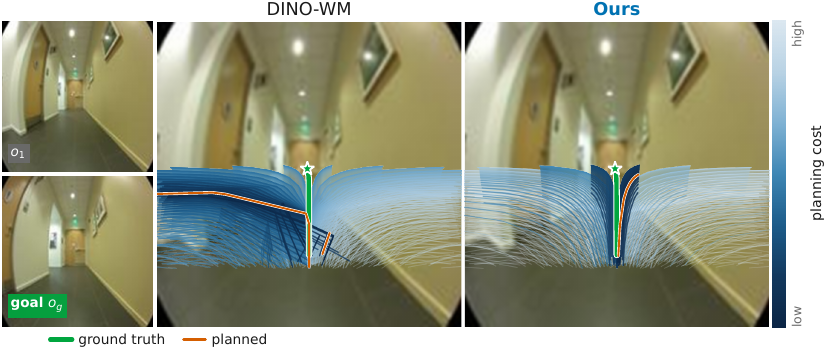}\\[3pt]
    \includegraphics[width=\linewidth]{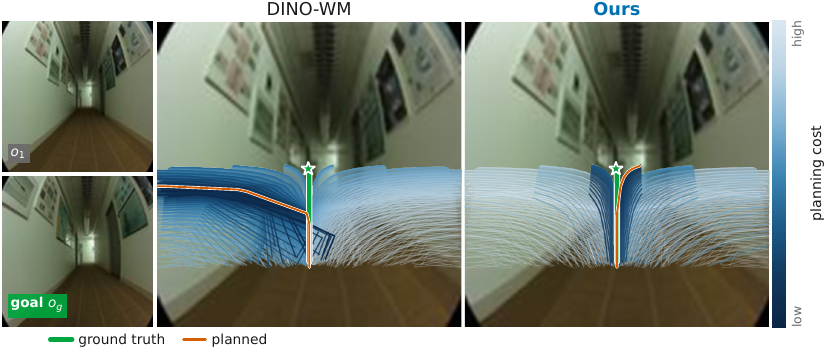}
    \vspace{-1mm}
    \caption{Additional CEM candidate visualizations on held-out GNM episodes (outdoor, top; indoor, below); legend as in Fig.~\ref{fig:gnm_candidates}.}
    \label{fig:gnm_candidates_more}
\end{figure}

Fig.~\ref{fig:gnm_candidates_more} shows five further held-out GNM episodes in the format of Fig.~\ref{fig:gnm_candidates}:
an outdoor dirt road (top) and four indoor corridors. The pattern matches the
main text across all of them: DINO-WM's cost has no basin around the
ground-truth trajectory, so its plan drifts, whereas our rollout+MCR model
forms a low-cost basin around the ground-truth trajectory.

\end{document}